\documentclass[10pt,twocolumn,letterpaper]{article}

\usepackage{iccv}
\usepackage{times}
\usepackage{epsfig}
\usepackage{graphicx}
\usepackage[font=small]{caption}
\usepackage{amsmath}
\usepackage{amssymb}
\usepackage{float}

\usepackage[table,dvipsnames]{xcolor}
\usepackage{booktabs}

\usepackage{url}

\usepackage[absolute]{textpos}
\def\ie{i.e.\ }
\def\eg{e.g.\ }

\usepackage{tabularx}
\newcolumntype{L}[1]{>{\raggedright\arraybackslash}m{#1}}
\newcolumntype{C}[1]{>{\centering\arraybackslash}m{#1}}
\newcolumntype{R}[1]{>{\raggedleft\arraybackslash}m{#1}}

\renewcommand{\emph}[1]{\textit{#1}}

\usepackage[pagebackref=true,breaklinks=true,letterpaper=true,colorlinks,bookmarks=false]{hyperref}

\usepackage[capitalize]{cleveref}
\crefname{section}{Sec.}{Secs.}
\Crefname{section}{Section}{Sections}
\Crefname{table}{Table}{Tables}
\crefname{table}{Tab.}{Tabs.}

\iccvfinalcopy 

\def\iccvPaperID{29} 

\ificcvfinal\fi

\begin{document}

\begin{textblock}{15}(3,25.8)
{\footnotesize\noindent\color{gray} © 2023 IEEE.  Personal use of this material is permitted.  Permission from IEEE must be obtained for all other uses, in any current or future media, including reprinting/republishing this material for advertising or promotional purposes, creating new collective works, for resale or redistribution to servers or lists, or reuse of any copyrighted component of this work in other works. The final version of this record is available at \url{https://doi.org/10.1109/ICCVW60793.2023.00460}.}
\end{textblock}

\title{Efficient 3D Reconstruction, Streaming and Visualization of Static and Dynamic Scene Parts for Multi-client Live-telepresence in Large-scale Environments}

\author{Leif Van Holland\textsuperscript{1}\quad Patrick Stotko\textsuperscript{1}\quad Stefan Krumpen\textsuperscript{1}\quad Reinhard Klein\textsuperscript{1}\quad Michael Weinmann\textsuperscript{1,2}\\
\\
\textsuperscript{1}University of Bonn\qquad \textsuperscript{2}Delft University of Technology
}

\twocolumn[{%
\renewcommand\twocolumn[1][]{#1}%
\maketitle
\begin{center}
    \centering
    \captionsetup{type=figure}
    \includegraphics[width=\linewidth]{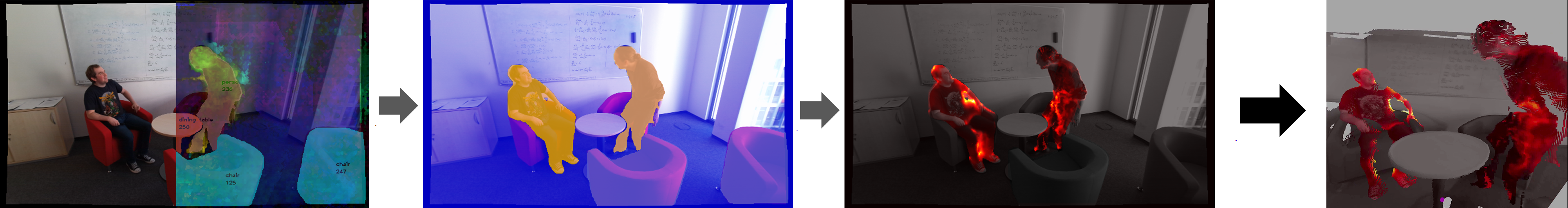}
    \captionof{figure}{Visualization of the key components of our proposed pipeline. The color image is blended with class and instance information, and shown along with the optical flow with respect to the previous frame (first image). This information is integrated to produce a mask that segments the frame into static and dynamic regions (second image). Together with an accumulated 3D motion estimate (third image), the scene is streamed to one or multiple remote clients for immersive exploration in VR (fourth image). In this example, the user chose to view the accumulated 3D motion.}
    \label{fig:teaser}
\end{center}%
}]

\ificcvfinal\thispagestyle{empty}\fi

\begin{abstract}
Despite the impressive progress of telepresence systems for room-scale scenes with static and dynamic scene entities, expanding their capabilities to scenarios with larger dynamic environments beyond a fixed size of a few square-meters remains challenging.

In this paper, we aim at sharing 3D live-telepresence experiences in large-scale environments beyond room scale with both static and dynamic scene entities at practical bandwidth requirements only based on light-weight scene capture with a single moving consumer-grade RGB-D camera.
To this end, we present a system which is built upon a novel hybrid volumetric scene representation in terms of the combination of a voxel-based scene representation for the static contents, that not only stores the reconstructed surface geometry but also contains information about the object semantics as well as their accumulated dynamic movement over time, and a point-cloud-based representation for dynamic scene parts, where the respective separation from static parts is achieved based on semantic and instance information extracted for the input frames.
With an independent yet simultaneous streaming of both static and dynamic content, where we seamlessly integrate potentially moving but currently static scene entities in the static model until they are becoming dynamic again, as well as the fusion of static and dynamic data at the remote client, our system is able to achieve VR-based live-telepresence at close to real-time rates.
Our evaluation demonstrates the potential of our novel approach in terms of visual quality, performance, and ablation studies regarding involved design choices. 
\end{abstract}


\section{Introduction}

Sharing immersive, full 3D experiences with remote users, while allowing them to explore the respectively shared places or environments individually and independently of the sensor configuration, represents a core element of \emph{metaverse} technology.
Beyond pure 2D images or 2D videos, 3D telepresence is defined as the impression of individually \emph{being there} in an environment that may differ from the user's actual physical environment~\cite{minsky1980telepresence,Fontaine:1992,held1992telepresence,Witmer:1998,draper1998telepresence}.
This offers new opportunities for diverse applications including remote collaboration, entertainment, advertisement, teaching, hazard site exploration, rehabilitation as well as for joining virtual sports events, work meetings, remote inspection, monitoring and maintenance, consulting applications or simply enjoying social gatherings.
In turn, the possibilities for virtually bringing people or experts together from all over the world in a digital twin of a location as well as the live-virtualization of such environments and events may reduce the effort regarding on-site traveling for many people, which not only helps to reduce our $CO_2$ footprint and increase the efficiency of various processes due to time savings, but also facilitates economically less well-situated or handicapped people to access such environments or events.

The creation of an immersive telepresence experience relies on various factors.
Respective core features are visually convincing depictions of a scenario as well as the subjective experience, vividness, and interactivity in terms of operating in the scene~\cite{steuer1992defining,schloerb1995quantitative}.
Therefore, the involved aspects include display parameters (e.g., resolution, frame rate, contrast, etc.), the presentation of the underlying data, its consistency, low-latency control to avoid motion sickness, the degree of awareness and the suitability of controller devices~\cite{minsky1980telepresence,Fontaine:1992,held1992telepresence,Witmer:1998,draper1998telepresence,steuer1992defining,schloerb1995quantitative}.
Furthermore, experiencing 3D depth cues like stereopsis, motion parallax, and natural scale also contribute to the perceived level of immersion and copresence~\cite{gibbs1999teleport,muhlbach1995telepresence}.

However, such immersive 3D scene exploration experience becomes particularly challenging for telepresence in live-captured environments due to the additional requirement of accurately reconstructing the digital twin of the underlying scene on the fly as well as its efficient streaming and visualization to remote users under the constraints imposed by available network bandwidth and client-side compute hardware.
Among many approaches, impressive immersive AR/VR-based live-3D-telepresence experiences have only been achieved based on advanced RGB-D acquisition for dynamic scenes on the scale of rooms, \ie areas of only a few square-meters, using special expensive static capture setups~\cite{gross2003blue,Vasudevan:2011,Maimone:2011,Maimone:2012,collet2015high,roberts2015withyou,Fairchild:2016,zioulis20163d,Orts-Escolano:2016,joachimczak2017real,komiyama2017jackin,su2017rgb,du2018montage4d,parikh2018mixed,cordova2019low,lawrence-2019-starline} and display technology~\cite{lawrence-2019-starline}, as well as for static scenes beyond that scale based on low-cost and light-weight incremental scene capture with a moving depth camera~\cite{mossel,Stotko2019SLAMCast:Telepresence,stotko2019ismar,stotko2019iros,stotko2019cvprw}.
For the latter category, bandwidth requirements have been reduced from hundreds of MBit/s for a single user~\cite{mossel} to around 15MBit/s for group-scale sharing of telepresence in live-captured environments while also handling network interruptions~\cite{Stotko2019SLAMCast:Telepresence,stotko2019ismar,stotko2019cvprw}, thereby even allowing live-teleoperation of robots~\cite{stotko2019iros}.
However, expanding the capabilities and, thereby, overcoming the aforementioned limitations in large dynamic environments for many users with low-cost setups still remains an open challenge.

In this paper, we aim at sharing 3D live-telepresence experiences in large-scale environments beyond room scale with \emph{both static and dynamic scene entities} at practical bandwidth requirements and based on light-weight scene capture with a single moving consumer-grade RGB-D camera.
For this purpose, we propose a respective system that relies on efficient 3D reconstruction, streaming and immersive visualization for dynamic large-scale scenes.

In particular, the key contributions of our work are:
\begin{itemize}
    \item For the sake of efficiency, our system leverages a hybrid volumetric scene representation, where we use optical flow and instance information extracted from the input frames to detect static and dynamic scene entities, thereby allowing the combination of a classic implicit surface geometry representation enriched with the object semantics as well as their accumulated dynamic motion over time, with a point-cloud-based representation of dynamic parts.
    \item We achieve efficient data streaming to remote users by the separate yet simultaneous streaming of both static and dynamics scene information, where we seamlessly integrate potentially moving but currently static scene entities in the static model until they are becoming dynamic again. Additionally, the fusion of static and dynamic data at the remote client allows VR-based visualization of the scene at close to real-time rates.
    \item We demonstrate the potential of our approach in the scope of several experiments and provide an ablation study for respective design choices.
\end{itemize}
Furthermore, while not being among the main contributions of our work, our approach also inherits the robustness of previous techniques to network interruptions for the reconstruction of the static scene parts as well as the scalability to group-scale telepresence~\cite{Stotko2019SLAMCast:Telepresence,stotko2019ismar,stotko2019iros}.
An overview of our proposed system is depicted in \Cref{fig:teaser}.

\section{Related Work}

\paragraph{Telepresence Systems}
Despite almost two decades of progress, the development of systems that allow immersive telepresence experiences remains challenging due to the prerequisite of simultaneously achieving high-fidelity real-time 3D scene reconstruction, the efficient streaming and management of the reconstructed models and the high-quality visualization based on AR and VR equipment.
Early approaches were limited by the capabilities of the available hardware~\cite{Fuchs:1994,Kanade:1997,Mulligan:2000,Towles:2002,Tanikawa:2005,Kurillo:2008} or inaccurate silhouette-based reconstruction techniques~\cite{Petit:2010,Loop:2013}.
Depth-based 3D scanning led to improved reconstruction quality and allowed telepresence at the scale of rooms~\cite{izadi2011kinectfusion,Maimone:2012,Maimone:2012b,Molyneaux:2012,Jones:2014,Fuchs:2014}, however, remaining artifacts induced by the high sensor noise and temporal inconsistency in the reconstruction process still impacted the visual experience.
More recently, advances in 3D scene capture, streaming, and visualization technology led to impressive immersive AR/VR-based live 3D telepresence experiences.
Live-telepresence for small-scale scenarios of a few square-meters has been achieved based on light-weight capture setups for teleconferencing~\cite{nguyen2014item,jones2009achieving,edelmann2012face2face,pejsa2016room2room,bell2019holo,cho2020effects} and other collaborative scenarios~\cite{zhang2012tele,Sodhi:2013,lu2015immersive,greenwald2019electrovr,teo2019mixed,fadzli2020robust} as well as based on expensive multi-camera static and pre-calibrated capture setups~\cite{gross2003blue,Vasudevan:2011,Maimone:2011,Maimone:2012,collet2015high,roberts2015withyou,Fairchild:2016,zioulis20163d,Orts-Escolano:2016,joachimczak2017real,komiyama2017jackin,su2017rgb,du2018montage4d,parikh2018mixed,cordova2019low,lawrence-2019-starline}.
Furthermore, live-telepresence for scenarios beyond a few square-meters has been achieved based on low-cost and light-weight incremental scene capture with a moving depth camera ~\cite{Bruder:2014,mossel,Stotko2019SLAMCast:Telepresence,stotko2019iros,stotko2019ismar,stotko2019cvprw,young2020mobileportation}, allowing remote users to immersively explore a live-captured environment independent of the sensor configurations. 
Regarding the latter approaches, impractical bandwidth requirements of up to 175 MBit/s for immersive scene exploration by a single user ~\cite{mossel} have been overcome by more recent approaches that allow group-scale sharing of telepresence experiences in live-captured environments while also handling network interruptions~\cite{Stotko2019SLAMCast:Telepresence,stotko2019ismar,stotko2019iros,stotko2019cvprw} as well as live-teleoperation of robots~\cite{stotko2019iros}.
Furthermore, mechanisms for annotation, distance measurement ~\cite{stotko2019iros} and efficient collaborative VR-based 3D labeling were added~\cite{zingsheim2021collaborative}.
However, practical sharing of live-captured 3D experiences in dynamic large-scale environments for many users with low-cost setup still remains an open challenge.
The same applies for immersive robot teleoperation, where approaches focused on small-scale scenarios with dynamics~\cite{peppoloni2015immersive,krupke2018prototyping,lipton2017baxter,theofanidis2017varm,whitney2018ros,rosen2020mixed,naceri2021vicarios} and large-scale, static scenarios~\cite{stotko2019iros}.

In contrast to the aforementioned approaches, we propose a live-telepresence system for large-scale environments while also taking scene dynamics into account. 

\paragraph{3D Reconstruction and SLAM Techniques}

Current state-of-the-art telepresence systems rely on depth-based simultaneous localization and mapping (SLAM) techniques.
Examples are the use of depth-sensor-based 3D scene capture based on surfels~\cite{henry2014rgb} or extensions of KinectFusion~\cite{Newcombe2011KinectFusion:Tracking,izadi2011kinectfusion} in terms of voxel block hashing techniques~\cite{niessner2013real,kahler2015very,kahler2015hierarchical,kahler2016real,prisacariu2017infinitam} for incremental scene capture for large-scale telepresence applications~\cite{mossel,Stotko2019SLAMCast:Telepresence,stotko2019ismar,stotko2019iros,stotko2019cvprw}.
To avoid the need for depth sensors, more recent SLAM approaches for incremental scene capture -- that might be applicable in respective telepresence applications -- leveraged principles of deep learning ~\cite{kuznietsov2017semi,yang2018deep,kendall2017uncertainties,klodt2018supervising,yang2020d3vo,czarnowski2020deepfactors,wimbauer2021monorec}.
Further approaches investigated 3D reconstruction from multiple synchronized cameras~\cite{moore2010synchronization,alexiadis2013real,duckworth2011camera,alexiadis2014fast,islam2014fast}.

Recently, neural scene representation and rendering techniques~\cite{tewari2020state,tewari2022advances} have led to significant improvements in reconstruction quality for small-scale objects or scenes.
The underlying idea originates from novel view synthesis and consists of training a neural network to represent a scene with its weights, so that respectively synthesized views match the input photographs.
In particular, this includes implicit scene representations based on Neural Radiance Fields (NeRFs)~\cite{mildenhall2021nerf} and respective extensions towards speeding up model training~\cite{reiser2021kilonerf,fridovich2022plenoxels,chen2022mobilenerf,deng2022depth,chen2022tensorf,wang2022r2l,sun2022direct,mueller2022instantngp,Fang2022,Cao2022rtNeRF,Chen2022tensorrf,Zhang2023Nerflets,Yariv2023baked,Mubarik2023HardwareAcc} with training times of seconds, the adaptation to unconstrained image collections~\cite{martin2021nerf,chen2022hallucinated,Seong2022hdrplenoxels}, deformable scenes~\cite{park2021nerfies,pumarola2021d,gafni2021dynamic,tretschk2021non,raj2021,noguchi2021neural,tseng2022cla,peng2021animatable,park2021hypernerf,chen2021animatable,liu2021neural,Jiang2022alignnerf,Li2022DynIBaR,Fang2022} and video inputs~\cite{li2021neural,xian2021space,du2021neural,peng2021neural,gao2021dynamic,li2022neural,Tancik2022blocknerf,Li2022streaming}, the refinement or complete estimation of camera pose parameters for the input images ~\cite{yen2021inerf,wang2021nerf,sucar2021imap,Chung-2022-Orbeez-SLAM,zhu2022nice,zhu2023nicer,rosinol2022nerf,zhang2022nerfusion,meng2021gnerf,lin2021barf,jeong2021self,Xia-2022-SiNeRF,Maggio-2023-Loc-NeRF,Bian-2022-NoPe-NeRF,Cheng20223LUNeRF,chen-2023-DBARF,Chen-2023-L2G-NeRF,Heo-2023-CamPos_MResHashEncoding,Sun2023NeRF-Loc,Liu2023NeRF-Loc}, combining NeRFs with semantics regarding objects in the scene ~\cite{vora2021nesf,zhi2021place,fu2022panoptic}, incorporating depth cues~\cite{wei2021nerfingmvs,deng2022depth,roessle2022dense,rematas2022urban,attal2021torf} to guide the training and allow handling textureless regions, handling large-scale scenarios~\cite{tancik2022block,turki2022mega,Mi2023switchnerf}, and streamable representations~\cite{cho2022streamable,Takikawa2022VBNF}.
However, despite promising results, current solutions~\cite{sucar2021imap,zhu2022nice,zhu2023nicer,rosinol2022nerf,zhang2022nerfusion,Maggio-2023-Loc-NeRF} do not yet reach real-time performance but only reach 12 FPS on a high-end GPU \cite{rosinol2022nerf} or less within completely static environments. Further improvements regarding efficiency and the handling of dynamic scenes are required to achieve real-time performance for the joint camera pose estimation and neural scene reconstruction in a SLAM setting in dynamic environments.

Particularly addressing dynamic environments, various approaches focused on filtering dynamic objects and only reconstructing the static background~\cite{kim2016effective,scona2018staticfusion,fan-2019,bescos2018dynaslam,yu2018ds,zhang2020flowfusion,xu2019mid,Xiao2019DynamicSLAM,Cui2019sofslam,Li2021dpslam,Wu2022yoloslam,Fan2022BlitzSLAM} or additionally reconstructing the dynamics based on rigid object tracking and reconstruction~\cite{stuckler2015efficient,wulff2017optical,li2017rgb,runz2017co,runz2018maskfusion,Henein:2020} and non-rigid object tracking and reconstruction~\cite{li2012temporally,keller2013real,ye2014real,newcombe2015dynamicfusion,guo2015robust,whelan2016elasticfusion,innmann2016volumedeform,whelan2016elasticfusion,dou2016fusion4d,zhang2017mixedfusion,slavcheva2017killingfusion,jaimez2017fast,slavcheva2018sobolevfusion,strecke2019fusion,Tretschk2000demea,Cai2022NeurIPS}.
Taking inspiration of the non-rigid scene reconstruction approaches in terms of separating static and dynamic scene parts, the 3D reconstruction approach involved in our live-telepresence system is particularly designed for capturing large-scale environments (i.e., beyond scenarios limited to a small area of a few square-meters) with both static and dynamic entities based on a single moved RGB-D camera. 
Our hybrid volumetric scene representation leverages semantic and instance information to detect dynamic scene entities and combines a voxel-based scene representation for the static parts, where we also accumulate information on whether and how significant objects have been moved, with a point-cloud-based representation of dynamic parts.
A major contribution of our work is the separate but simultaneous streaming of both static and dynamics scene information and its VR-based visualization at close to real-time rates.

\begin{figure*}[t]
\includegraphics[width=\textwidth]{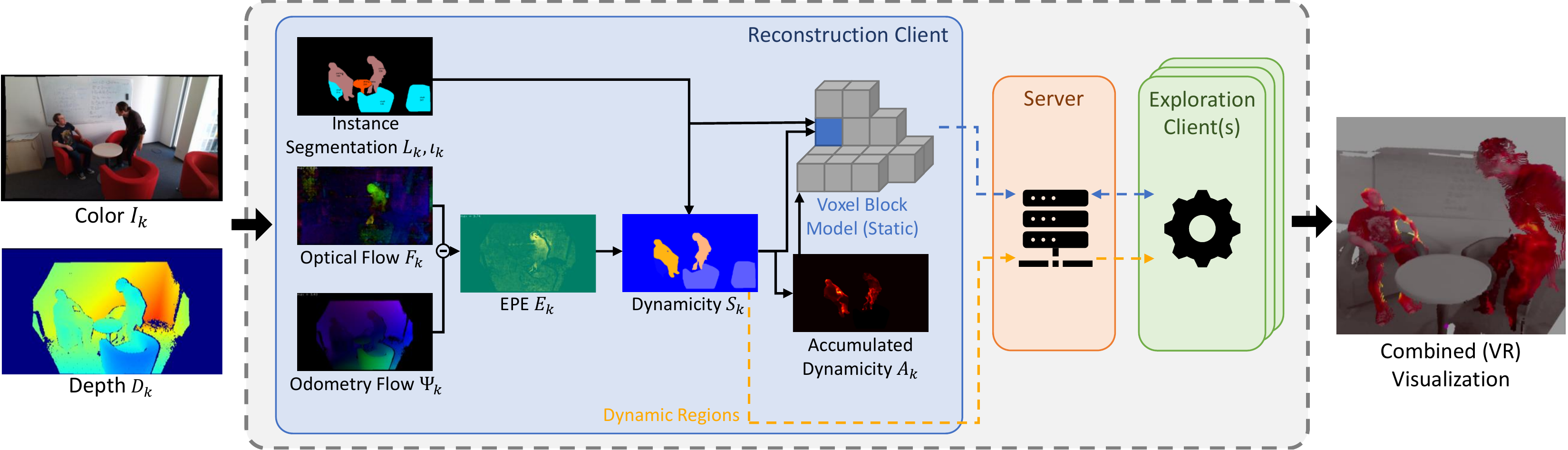}
\caption{
Visualization of different processing stages for the $k$-th RGB-D frame in the pipeline.
Starting with color $I_k$ and depth $D_k$, instance segmentation $L_k$ (class labels) and $\iota_k$ (instance IDs), optical flow $F_k$ and odometry flow $\Psi_k$ (i.e., the flow generated from the estimated camera motion) are computed.
Next, the end-point-errors (EPE) between the flows are computed, normalized and propagated using the instance segmentation to generate the dynamicity scores $S_k$. The scores are accumulated in $A_k$ and $L_k, \iota_k, S_k$ and $A_k$ are used to integrate information about static regions in the voxel block model.
New static voxels and current dynamic regions are sent to the server, which forwards this information to the exploration clients appropriately.
}
\label{fig:pipeline}
\end{figure*}

\section{Methodology}

As shown in Figure~\ref{fig:pipeline}, our live-telepresence system for large-scale environments with scene dynamics at practical bandwidth requirements takes a continuous stream of RGB-D images $(I_1, D_1), (I_2, D_2), ...$ from a moving depth camera as input, where
${I_k(u)\in\mathbb{R}^3}$ represents the red, green, and blue color values of frame $k$, and ${D_k(u)\in\mathbb{R}}$ the corresponding raw depth measurement at pixel ${u\in\mathcal{U}\subset\mathbb{N}^2}$, with $\mathcal{U}$ being the image domain.
The main challenge consists in an efficient processing of these measurements, their efficient integration into a consistent model and the efficient streaming of the latter over the network at practical bandwidth requirements to remote clients, where it has to be visualized at adequate visual quality and at tolerable overall latency.
For this purpose, we use a hybrid scene representation that separately handles static and dynamic scene parts, thereby allowing the combination of efficient large-scale 3D scene mapping techniques, that face problems with dynamic regions, with efficient point-based reconstruction for the dynamic parts.
In more detail, we segment the frames of the input stream into static and dynamic regions by determining score maps ${S_k}$, where ${S_k(u)\in\mathbb{R}}$ describes the amount of dynamicity in frame $k$ at pixel $u$.
This separation allows us to efficiently reconstruct, stream and immersively visualize static regions using existing state-of-the-art large-scale telepresence techniques~\cite{Stotko2019SLAMCast:Telepresence,stotko2019ismar} while simultaneously reconstructing, streaming and visualizing dynamic scene parts based on a point-based representation in terms of a partial \mbox{RGB-D} image and its corresponding estimated camera pose, thereby limiting the amount of data to be transferred and reducing the processing time.
After streaming the hybrid scene representation to remote users, its static and dynamic parts are joined in a combined 3D visualization.
In the following subsections, we explain the different steps of our pipeline.
Please refer to the supplemental material for more details.

\subsection{Segmentation into Static and Dynamic Regions}
\label{sec:static_dynamic}

For the sake of efficiency, we segment the RGB-D frames of the input stream into static and dynamic regions, which will later allow the efficient treatment of the different types of scene parts.
For this purpose, we compute the aforementioned score maps $S_k$.
In the following, we will ensure that these scores are normalized in the sense that a pixel is deemed static if ${S_k(u) \leq 1}$, and dynamic if ${S_k(u) > \tau}$, where ${\tau \geq 1}$ is a threshold that allows for a region of uncertainty between the static and dynamic labels.
\paragraph{Instance Segmentation}
To compute the dynamicity score $S_k$ of frame $k$, we first detect objects in $I_k$ using instance segmentation \cite{yolov8}, which yields both a class label and an instance ID for each pixel in the image, \ie ${(L_k, \iota_k) = f_\text{seg}(I_k)}$ of $I_k$, where ${L_k(u)\in\mathbb{N}}$ is the predicted class label and ${\iota_k(u)\in\mathbb{N}}$ is the instance ID at pixel $u$.
The raw output of the segmentation network may consist of multiple, potentially overlapping region proposals, which we integrate into the instance and label maps using non-maximum suppression. 
The resulting indices are then associated with the IDs from the previous frame to get the final map $\iota_k$ of instance IDs.
The label map $L_k$ is set to the class labels corresponding to the instances.
In our experiments, we used YOLOv8 \cite{yolov8} as the segmentation network.
\paragraph{Optical Flow Estimation}
Next, we estimate the backward optical flow ${F_k = f_\text{flow}(I_k, I_{k-1})}$, where ${F_k(u)\in\mathbb{R}^2}$ is the corresponding flow vector at pixel $u$, such that $u$ in $I_k$ corresponds to ${u + F_k(u)}$ in $I_{k-1}$.
For $f_\text{flow}$, we use the NVIDIA Optical Flow Accelerator (NVOFA) \cite{nvofa}.
We additionally generate a map of confidence weights $W_k(u)\in[0,1]$ based on the agreement with the inverse flow and per-pixel costs given by the NVOFA to reduce the influence of bad correspondences.
\paragraph{Odometry}
Subsequently, we estimate the camera motion $\xi_k\in\mathfrak{se}(3)$ between the previous and current frame, yielding an absolute camera pose ${T_k\in\mathbb{R}^{4\times 4}}$ when we assume $T_1$ to be centered at the world origin.
Our implementation uses a standard point-to-plane RGB-D registration implementation \cite{Zhou2018open3d}.

\paragraph{End-point-error}
Based on $F_k$, $T_k$ and ${W_k}$, we determine a per-pixel end-point-error $E_k$ between the estimated flow and the flow $\Psi_k$ we expect from a completely static scene where only the camera is moving by $\xi_k$, i.e.,
\begin{equation}
    E_k(u) = W_k(u)\cdot\lVert F_k(u) - \Psi_k(u)\rVert_2.
\end{equation}
\paragraph{Dynamicity Score}
\label{par:dynamicity_score}
To decide which of the resulting scores $E_k(u)$ indicate dynamic regions, we found that a simple thresholding is not sufficient, because the average error varies too strongly, especially for frame pairs where the camera tracking or optical flow network yield poor estimates.
To reduce the influence of these fluctuations, we instead analyze the histogram of per-pixel errors for all pixels of each instance $i$.
More specifically, we are selecting the rightmost mode $s_k(i)$ that is above a minimum size threshold by finding the corresponding bin index and choosing $s_k(i)$ as the center of that bin.
We normalize all scores by subtracting the smallest mode from them, assuming that at least one of the detections is of static nature.
Together with an empirically chosen linear rescaling by a factor ${\delta\in\mathbb{R}_{\geq 0}}$, we get the normalized scores
\begin{equation}
    E'_k(u) = \delta \cdot (E_k(u) - \min_i\{s_k(i)\}),
\label{eq:error_normalization}
\end{equation}
which fulfill the previously mentioned criterion that scores $S_k(u)\leq 1$ are indicating a static object, while higher scores indicate dynamic regions.
While $E'_k(u)$ can now be used for the segmentation into static and dynamic regions, we found the visualization of moving regions to be more coherent if the segmentation happens on the object level.
This is particularly important for articulated or non-rigid objects like humans, where potentially only a small part of the object (\eg an arm) is moving.
To accomplish this, we use the normalized modes $s'_k(i)$, which result from applying the transformation from \Cref{eq:error_normalization} to $s_k(i)$.
An instance $i$ is deemed as dynamic if its normalized mode is above the dynamic threshold $\tau$, i.e., ${s'_k(i) \geq \tau}$.
To represent this in the resulting score map, we propagate $s'_k(i)$ in the final score map by setting ${S_k(u) = s'_k(i)}$ for all pixels $u$ with ${\iota_k(u)=i}$, i.e., all pixels belonging to instance $i$.
The score of each instance is temporally smoothed to be more robust against outliers.

\paragraph{Score Accumulation}
As the object tracking is only performed in 2D for efficiency reasons, we also accumulate the dynamicity scores of each instance over time in 2D by updating an accumulation map ${A_k(u)\in\mathbb{R}_{\geq 0}}$.
To increase the interpretability of the scores, we compute a 3D end-point-error between the last and current frame by using $F_k$ for the correspondences between the pixels and backprojecting the respective coordinates of into 3D using the corresponding depth maps and camera poses.
The resulting 3D flow ${\hat{F}_k(u)\in\mathbb{R}^3}$ is then combined with the warped previous accumulated score $A'_{k-1}$ to ${A_k(u) = A'_{k-1}(u) + \lVert\hat{F_k}(u)\rVert_2}$.
We also experimented to use ${\hat{F}_k}$ as an input for the end-point-error calculation, but found that the signal was too noisy for our method to robustly distinguish between static and dynamic scene parts.

\subsection{Update of the Static Model}
\label{sec:static_model_update}

With the score map $S_k$ computed, we are able to integrate the static part of the frame into the static model.
For this purpose, we use a modified version of real-time 3D reconstruction based on spatial voxel block hashing~\cite{niessner2013real}, with an added extension for concurrent retrieval, insertion, and removal of data~\cite{Stotko2019SLAMCast:Telepresence}.
However, in order to further increase the efficiency of the approach, we seamlessly shift potentially dynamic but currently static scene parts into the static scene representation until they become dynamic again. This requires us to additionally consider the following situations:

\begin{enumerate}
    \item \label{situation_dynamic} Dynamic regions should not be integrated into the static model. In case this happens erroneously, they should be removed as quickly as possible.
    \item \label{situation_dynamic_to_static} Regions that change their state from dynamic to static (\eg a box was placed on a table) should be integrated into the static model seamlessly.
    \item \label{situation_static_to_dynamic} Regions changing their state from static to dynamic (\eg a box is picked up) should be removed from the static model immediately.
    \item \label{situation_static_oof} Static regions that changed while not in the camera frustum should be updated as soon as new information is available.
\end{enumerate}

Following the suggested modification of the weighting schema for dynamic object motion by Newcombe et al. \cite{Newcombe2011KinectFusion:Tracking}, we truncate the updated weight, which effectively results in a moving average favoring newer measurements.
We extended the schema to incorporate the previously computed dynamicity scores.
This helps in situations \ref{situation_dynamic} and \ref{situation_static_to_dynamic}, since dynamic regions are updated with new information more quickly, as well as in situation \ref{situation_static_oof}, as the weight is truncated even for static regions.
In addition, we aid the timely removal of dynamic regions from the static model (situations \ref{situation_dynamic} and \ref{situation_static_to_dynamic}) by setting the SDF value to $-1$ for voxels where the associated dynamicity score $S_k(u)$ exceeds a threshold ${\tau_\text{SDF}>0}$.
In conjunction with the modified integration weight, this invalidates the existing surface estimate at that location.
Situation \ref{situation_dynamic_to_static} is covered by the temporal smoothing of the scores.
Details can be found in the supplementary material.
\subsection{Visualization}
After having streamed the hybrid scene representation to remote users' devices, the static and dynamic scene entities have to be combined within an immersive scene exploration component, where we focus on VR-based immersion of users into the live-captured scenarios.
For this, we created a client component that receives updates of the static model as well as the dynamic regions of the current RGB-D frame.

The static model is visualized as a mesh, where the local mesh representation of the static scene is updated using received MC voxel block indices and rendered in real time, thereby following previous work~\cite{Stotko2019SLAMCast:Telepresence}.
In contrast, the dynamic parts are shown as a point cloud at the corresponding location relative to the static mesh.
For this, we backproject the dynamic pixels of the current RGB-D frame using the known camera intrinsics and the current camera pose.

The user is then able to individually and independently of the sensor explore the captured scene by physically looking and walking around, or use a teleportation functionality for locomotion.
The current position and orientation of the RGB-D sensor and other users is also shown.

\subsection{Streaming}

To be able to run the described method with low latency from the time of capturing to the visualization at remote locations, we use a server-client architecture.
The server receives and distributes data packages over a network to the appropriate processing clients.
The RGB-D capture, segmentation into static and dynamic regions as well as the integration into the static model are performed in the \emph{reconstruction client}.
Updates of this representation are then broadcasted to one or multiple \emph{exploration clients}, which in turn update a mesh representation of the static scene using the MC indices.
At the same time, the server also sends updates of the dynamic regions as masked RGB-D images together with the current camera pose estimate, such that the RGB-D pixels can be projected into the scene as a point cloud.
For all network communication, we use a general-purpose lossless data compression scheme \cite{zstandard} to reduce the bandwidth requirements.

\subsection{Implementation Details}

To take advantage of modern multiprocessor architectures, the stages shown in \Cref{fig:pipeline} are running concurrently, such that each stage can start with the next item once the processing of the current one has been completed.
While this leads to overhead due to inter-process communication, the processing speed of the pipeline is no longer bound to the latency, but the processing duration of the slowest stage in the pipeline.
We provide a more detailed, per-stage performance analysis in the supplemental material.

\section{Experimental Results}
To evaluate the performance of the proposed pipeline, we ran experiments on 10 self-recorded sequences captured with a Microsoft Azure Kinect RGB-D sensor in different office environments, and measured both speed and bandwidth metrics.

The scenes contain varying types of motion, and we categorized them into three groups.
Fixed (F.) are scenes that have no camera motion once dynamic entities can be seen in the camera, whereas Moving (M.) describes scenes with an always-moving camera and simultaneous object motion.
A third category Outside (O.) contains a scene where the camera is hand-held, but object motion only happens outside the camera view. 
A short description and some exemplary images of each scene are shown in the supplemental material.
To validate design choices, we also conducted an ablation study regarding certain components of the pipeline and compared them to baseline methods.
Following that, we will discuss the impact and limitations of the approach.

\subsection{Experimental Setup}
We set up three computers in a local network that each run one of the three processes shown in \Cref{fig:pipeline}.
All devices use the same hardware except for the GPU, which is an Nvidia GeForce RTX 3090 for the reconstruction client and an Nvidia GeForce GTX 1080 for both server and exploration client, as they require less GPU performance.
We measured three different metrics in this setup:
The end-to-end latency of an RGB-D frame from the camera to the exploration client, the frame-rate at which RGB-D frames are being processed by the components of the pipeline, and the network bandwidth between server and connected clients.
The latency and frame-rate is measured using timestamped logs that are synchronized between all computers to ensure a minimal deviation.
The frame-rate is given as the averaged arrival time difference between consecutive dynamic RGB-D images at the exploration client, and the latency is the average between the emission times of RGB-D frames into the pipeline and the corresponding arrival times at the exploration client.
The hyperparameters used for the performance evaluation and visualization were fixed for all scenes and are listed in the supplemental.

\subsection{Evaluation of Performance and Visual Quality}

\Cref{tab:latency_fps_results} shows the results of the frame-rate and latency measurements.
Here, the performance is largely independent of the type of scene and exhibits an average of around 0.4 seconds in end-to-end latency and a frame-rate of 18.8 frames per second (FPS).
A closer analysis reveals that the frame-rate is upper-bound by the single image inference speed of the instance segmentation network.
We refer to the supplemental for details.

\begin{table}
    \centering
\footnotesize
\begin{tabular}{c|c|c|c|c|c}
    \toprule
    Scene & F. & M. & O. & end-to-end [s] & FPS [1/s] \\
    \midrule
    items\_1  & \checkmark & & & 0.40 (0.02) & 18.95 (5.75) \\
    items\_2  & \checkmark & & & 0.40 (0.03) & 18.67 (5.95) \\
    people\_1 & \checkmark & & & 0.39 (0.02) & 19.33 (5.94) \\
    people\_2 & &\checkmark  & & 0.41 (0.03) & 17.92 (5.12) \\
    people\_3 & &\checkmark  & & 0.43 (0.03) & 17.73 (4.81) \\
    people\_4 & &\checkmark  & & 0.40 (0.03) & 17.98 (5.34) \\
    people\_5 & &\checkmark  & & 0.40 (0.02) & 19.01 (5.64) \\
    ego\_view & &\checkmark  & & 0.40 (0.02) & 18.93 (5.79) \\
    oof\_1    & & & \checkmark & 0.40 (0.08) & 19.81 (6.05) \\
    oof\_2    & & & \checkmark & 0.40 (0.02) & 19.56 (6.39) \\
    \bottomrule
\end{tabular}
    \caption{Performance results on the 10 self-recorded scenes.
    The F., M., O. columns indicate the type of motion that was captured (F: fixed camera when object motion is seen, M: camera always in motion, O: static scene manipulation outside of camera view).
    Latency and FPS columns show both the mean and standard deviation (in parentheses) of the respective metrics.}
    \label{tab:latency_fps_results}
\end{table}

\begin{table}
    \centering
\footnotesize
\begin{tabular}{c|c|c|c}
    \toprule
    Type & F. & M. & O.  \\
    \midrule
    TSDF & 44.80 (77.90) & 69.72 (92.14) & 65.70 (81.21) \\
    MC & \phantom{0}3.89 \phantom{0}(6.13) & \phantom{0}6.10 \phantom{0}(7.31) & \phantom{0}6.30 \phantom{0}(5.83) \\
    Dyn. & \phantom{0}7.06 \phantom{0}(7.37) & \phantom{0}7.41 \phantom{0}(8.43) & \phantom{0}2.66 \phantom{0}(4.42)\\
    \bottomrule
\end{tabular}
    \caption{Required mean bandwidth and respective standard deviation (in parentheses) in MBit/s of the different types of data packages over the types of recorded scenes (F: fixed camera when object motion is seen, M: camera always in motion, O: object motion only outside of camera view).
    }
    \label{tab:bandwidth_results}
\end{table}

\begin{figure*}
\centering
\includegraphics[width=\textwidth]{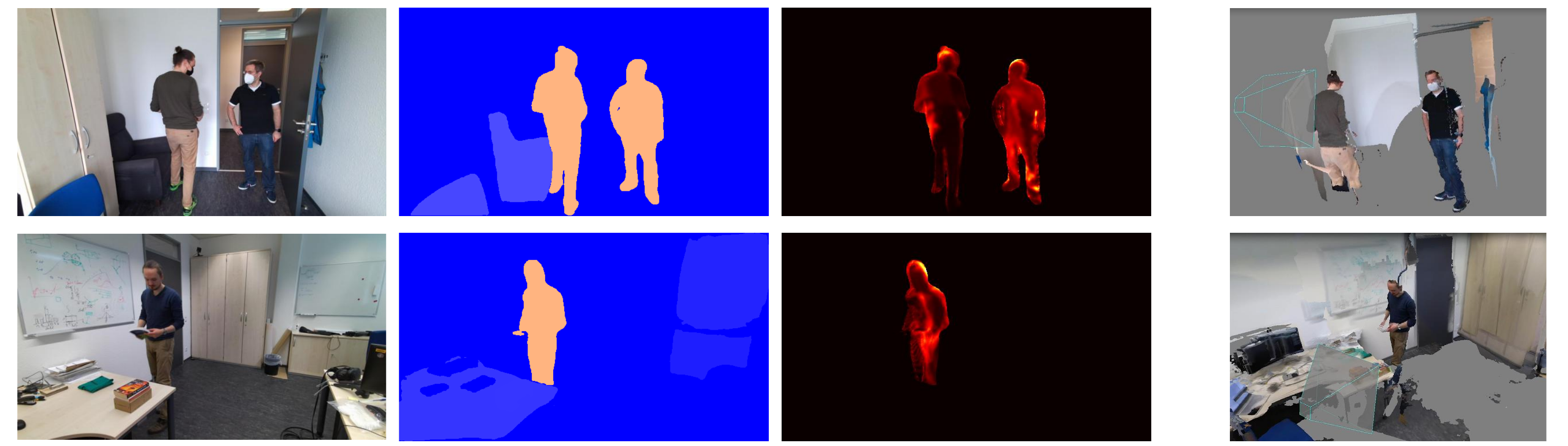}
\caption{Results of our approach on different scenes. Left to right: Input color image; resulting segmentation into static (blue) and dynamic (yellow) regions; the accumulated 3D flow magnitude; a novel view of the scene as visualized in the exploration client.}
\label{fig:visual_results}
\end{figure*}

The network bandwidth requirements are summarized in \Cref{tab:bandwidth_results}. 
Here, the measured package sizes are split up in the type of data.
TSDF represents the values of the truncated signed-distance function generated by the voxel block hashing of the reconstruction client, MC labels the Marching Cubes indices the server generates from the TSDF representation and sends to the exploration client(s).
The dynamic RGB-D that results from the segmentation of the reconstruction client and that is subsequently sent to the exploration client(s) is called Dyn.
The results indicate that the majority of data is transferred between reconstruction client and server.
The Marching Cubes indices and dynamic RGB-D data, which are selectively streamed to the exploration client(s), allow for multiple connections, even over the Internet, considering modern bandwidth availability.
Furthermore, we provide qualitative results in \Cref{fig:visual_results}.

\subsection{Ablation Study}

\begin{figure}
    \centering
    \includegraphics[width=\linewidth]{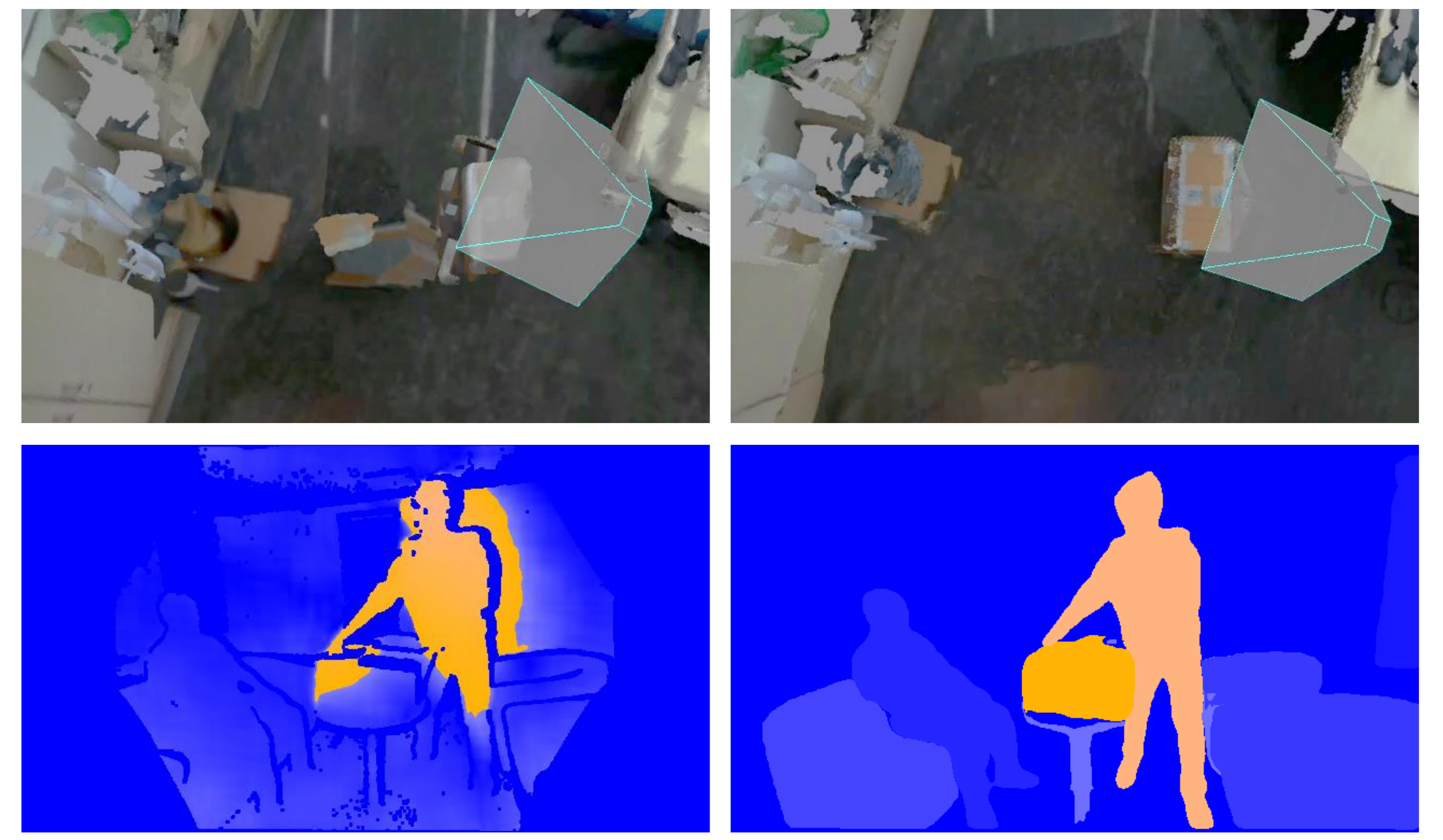}
    \caption{Comparison of design choices of the proposed pipeline. Top row: An example output from the exploration client using the standard voxel block weighting schema (left) vs. exponential weight decay via weight capping. The second approach yields a reconstruction of the box with fewer artifacts. Bottom row: Thresholding of the normalized EPE before (left) and after (right) propagation of the error modes into the static (blue) and dynamic (yellow) object masks. Again, the second approach produces a more plausible segmentation into static and dynamic regions.}
    \label{fig:ablation}
\end{figure}

To validate some design choices of our approach, we show the effects of removing certain elements of the pipeline on the results.
\Cref{fig:ablation} illustrates the effect of the weighting function from \Cref{sec:static_model_update}, as well as the difference between error thresholding with and without propagation into the object mask (\Cref{par:dynamicity_score}, \nameref{par:dynamicity_score}).
In the weighting example, we show that the update of inconsistent measurements results in less artifacts while walking around the box when using an exponential decay.
This motivates our choice to enable this weighting schema for regions with recent object motion.
At the same time, the floor texture shows slightly more artifacts as the more recent measurements are favored, but collide visually with regions that were not recently seen by the camera.
This effect is reduced in the original weighting schema, which motivates the extension of the schema mentioned in \Cref{sec:static_model_update}.

The bottom row of \Cref{fig:ablation} shows how the propagation of the error modes into the object masks aids to correctly identify potentially dynamic objects.
Due to weak motion boundaries produced by $f_\text{flow}$, a large region of pixels behind the moving person is considered dynamic after normalization.
This can be filtered out completely in this case using our approach.
We also conducted a performance comparison with different optical flow and instance segmentation approaches to validate our choice.
The results can be found in the supplemental material.
\subsection{Limitations}
\begin{figure}
    \centering
    \includegraphics[width=\linewidth]{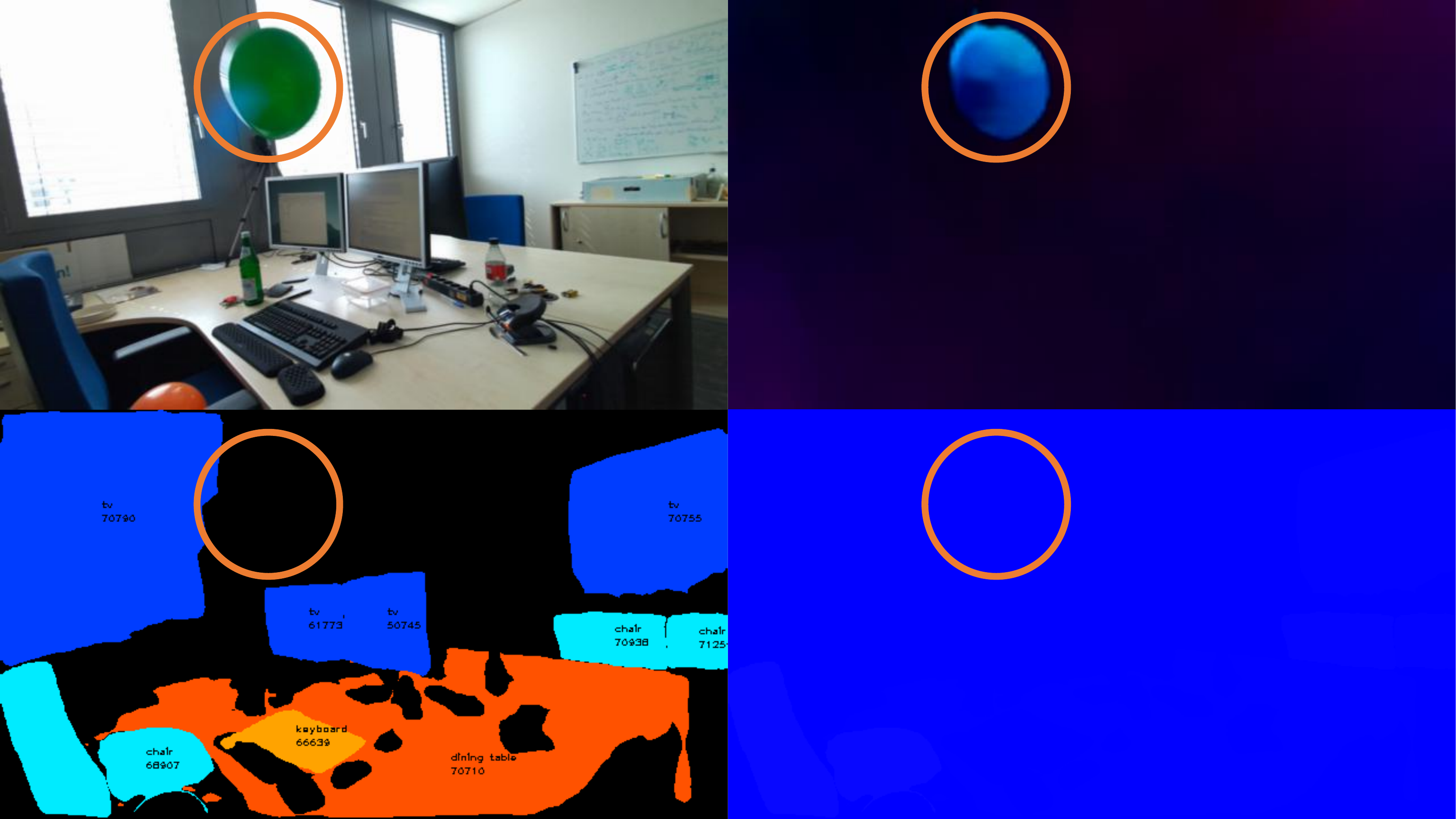}
    \caption{Failure case of our method. Shown are RGB (top left), optical flow (top right), instance segmentation (bottom left) and resulting segmentation into static and dynamic (bottom right). Even though a clear motion cue is available in the optical flow image, due to a missing object detection, our method fails to correctly identify the dynamic region (orange circle).}
    \label{fig:failure_case}
\end{figure}

While our approach shows promising results and is designed with modularity and extensibility in mind, there are also some limitations to consider.
Most importantly, the pipeline only runs at frame-rates close to real-time due to the performance limitations inherited by the involved neural network approaches.
In our scenario, we require high single-image inference speed, which is not a functionality most modern deep learning approaches are particularly tuned for.
Furthermore, our approach requires the segmentation network to detect objects to be able to identify dynamic regions, which limits its capabilities on out-of-distribution samples (\Cref{fig:failure_case}).
This is also the case for the optical flow network, as it is also limited by the quality of the training data and the domain overlap with the scenes we recorded.
However, due to the modular nature of our approach, future developments with improved accuracy of the predictions might address this current limitation of our approach. Furthermore, future developments on increasing the efficiency of the networks for the respectively involved subtasks will further improve the overall performance.

\section{Conclusions}

We presented a novel live-telepresence system that allows immersing remote users into live-captured environments with static and dynamic scene entities beyond an area of a few square-meters at practical bandwidth requirements.
In order to allow the respectively required efficient 3D reconstruction, data streaming and VR-based visualization, we built our system upon a novel hybrid volumetric scene representation that combines a voxel-based representation of static scene geometry enriched by additional information regarding object semantics as well as their accumulated dynamic movement over time with a point-cloud-based representation for dynamic parts, where we perform the respective separation of static and dynamic parts based on optical flow and instance information extracted for the input frames.
The separation, determined frame-by-frame on the 2D RGB-D data, remains unaffected by the length of the input sequence and scale of the scene and therefore does not impact the performance of the static reconstruction technique employed.
As a result of independently yet simultaneously streaming static and dynamic scene characteristics while keeping potentially moving but currently static scene entities in the static model as long as they remain static, as well as their fusion in the visualization on remote client hardware, we achieved VR-based live-telepresence in large-scale scenarios at close to real-time rates.
With the rapid improvements in hardware technology, particularly regarding GPUs, we expect our system to soon reach full real-time capability.
Also, the modularity of our system allows replacing individual components with newer approaches, which might be particularly relevant for the instance segmentation network as it represents the main bottleneck of our current system.
\section*{Acknowledgements}
This work was supported by the DFG project KL 1142/11-2 (DFG Research Unit FOR 2535 Anticipating Human Behavior) and NFDI4Culture (DFG Project Number 441958017).

{\small
\bibliographystyle{ieee_fullname}
\bibliography{literature}
}

\end{document}